\documentclass[a4paper,11pt]{amsart}

\usepackage[T1]{fontenc}    
\usepackage{textcomp}       
\usepackage[utf8]{inputenx} 

\DeclareFixedFont{\ttb}{T1}{txtt}{bx}{n}{8} 
\DeclareFixedFont{\ttm}{T1}{txtt}{m}{n}{8}  

\usepackage[hang]{caption}
\usepackage{url}
\usepackage{graphicx}

\usepackage{color}
\definecolor{deepblue}{rgb}{0,0,0.85}
\definecolor{deepred}{rgb}{0.6,0,0}
\definecolor{deepgreen}{rgb}{0,0.5,0}
\definecolor{deepbrown}{rgb}{0.88,0.52,0.26}
\definecolor{lavender}{rgb}{0.81,0.41,0.83}

\usepackage{listings}

\newcommand\pythonstyle{\lstset{
language=Python,
basicstyle=\ttm,
morekeywords={str},                
keywordstyle=\ttb\color{deepblue},
emph={MyClass,__init__},          
emphstyle=\ttb\color{deepred},    
stringstyle=\color{deepgreen},
commentstyle = \color{deepgreen},
frame=tb,                         
showstringspaces=false,
classoffset=2,
morekeywords={self},
keywordstyle=\color{deepbrown},
}}

\lstnewenvironment{python}[1][]
{
\pythonstyle
\lstset{#1}
}
{}

\newcommand\pythoninline[1]{{\pythonstyle\lstinline!#1!}}

\DeclareMathOperator*{\argmax}{argmax}

\title[Some approaches used to overcome overestimation]
{Some approaches used to overcome overestimation 
in Deep Reinforcement Learning  algorithms}         
         \author{Rafael Stekolshchik}

\begin{document}

\date{}

\begin{abstract}
Some phenomena related to statistical noise which have been investigated
by various authors under the framework of deep reinforcement learning (RL) algorithms are discussed.
The following algorithms are examined: the deep $Q$-network
(DQN), double DQN, deep deterministic policy gradient (DDPG),
twin-delayed DDPG (TD3), and hill climbing algorithm.
First, we consider {\it overestimation}, which is a harmful property resulting from noise.
Then we deal with noise used for {\it exploration}, this is the useful noise.
We discuss setting the noise parameter in the TD3 for typical PyBullet environments associated
with articulate bodies such as HopperBulletEnv and Walker2DBulletEnv.
In the appendix, in relation to the hill climbing algorithm,
another example related to noise is considered - an example of adaptive noise.
\end{abstract}

\maketitle

\section{\sc\bf Introduction}

In 1993, Thrun and Schwartz in \cite{1} gave an example in which overestimation
(caused by noise) asymptotically led to suboptimal policies.
On the other hand, adding noise to an action space helps algorithms more
efficiently perform exploration, which is not correlated with something unique,
see \cite{5}. We look at deep reinforcement learning (RL) algorithms
in terms of issues related to noise. In this article, we touch on the following algorithms:
the deep Q-network (DQN), double DQN, deep deterministic policy gradient (DDPG),
twin-delayed DDPG (TD3) and hill climbing algorithm.

In Section \ref{sec_overestim}, we present an overview of the approaches
that researchers have taken to overcome overestimation in models.
The first step is to decouple the action selection and action evaluation process.
This was realized in double DQN model. The second step relates to the actor-critic architecture;
here we decouple the value neural network (critic) from
the policy neural network (actor). In essence, this is a generalization of what
was done in the double DQN for the continuous action space case.
The DDPG and TD3 algorithms use this architecture, \cite{3, 4}.
A very significant advantage of the TD3 in overcoming overestimation
is the use of auto-critic with two-critic architecture.

In Section \ref{sec_explore}, we consider how exploration is implemented in the DQN,
double DQN, DDPG and TD3. Exploration is a major challenge when performing learning.
The main issue of Section \ref{sec_explore} is exploration noise.
Neural network models utilizing noise parameters have better exploration capabilities
and more successfully complete  deep RL algorithms.
A certain problem occurs when finding the true noise parameter for exploration.
We discuss the setting of this parameter in the TD3 for PyBullet environments
associated with articulate bodies such as HopperBulletEnv and Walker2DBulletEnv.

In Section \ref{sec_pybullet}, we will look at several experiments with PyBullet agents
related to articulated bodies: Hopper, Walker2D and HalfCheetah.  
In the spotlight aspects related to noise parameters.

In the Appendix, we consider hill climbing, a simple gradient-free algorithm.
This algorithm adds adaptive noise directly to the input variables, namely
to the weight matrix used for determining the neural network. 
\section{\sc\bf Efforts to overcome overestimation: Overview of approaches}
  \label{sec_overestim}
The DQN and double DQN algorithms
turned out to be very successful in the
cases involving discrete action spaces. However, it is known that these algorithms
suffer due to overestimation, see \cite{13}. This harmful property is much worse than
underestimation, because underestimation does not accumulate. Let us see
how researchers have tried to overcome overestimation.

\subsection{Overestimation in the DQN}

Let us consider the operator used for the calculation of the
target value $G_t$ in the key $Q$-learning equation (a.k.a the state-action-reward-state-action (SARSA) equation). This operator is called the maximization operator:
~\\
\begin{equation*}
  \label{DQN_key_eq_1}
    G_t = R_{t+1}  + \gamma\max_a Q(S_{t+1}, a) \\
\end{equation*}
~\\
Suppose, that the evaluation value for $Q(S_{t+1}, a)$ is already
overestimated. Then, the agent observes that error also accumulates for $Q(S_t, a)$:

\begin{equation*}
  \label{DQN_key_eq_2}
  \begin{split}
    & Q(S_t, a) \longleftarrow Q(S_t, a) + \alpha(R_{t+1} + \gamma\max_a Q(S_{t+1}, a) - Q(s_t, a_t)), \text{ or}\\
    & Q(S_t, a) \longleftarrow Q(S_t, a) + \alpha(G_t - Q(s_t, a_t)).
  \end{split}
\end{equation*}
~\\
~\\
Here, $R_t$ is the reward at time $t$, $G_t$ is the cumulative reward
and $Q(s, a)$ is the $Q$-value table of the shape  [space $\times$ action], \cite{11}.

In 1993, Thrun and Schwartz observed  that using function
approximators (i.e., neural networks) instead of simply utilizing lookup tables
(this is the basic technique of $Q$-learning) causes some noise in the output predictions.
They provided an example in which overestimation asymptotically leads to suboptimal policies,
see \cite{1}.

\subsection{Decoupling in the double DQN}

In 2015, Haselt et. al. showed that estimation errors can drive the obtained estimates up
and away from the true optimal values. They proposed a solution that reduces
overestimation in the discrete case: the double DQN, \cite{2}.
The important aspect of the double DQN is that it decouples
the action selection process from the action evaluation process.
Let us make this clear.

\begin{figure}[h]
\centering
\includegraphics[scale=1.2]{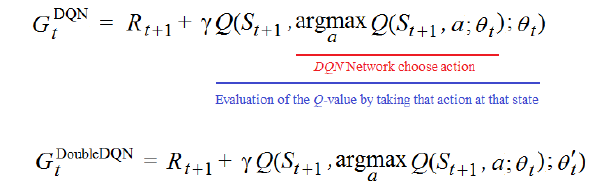}
\end{figure}

\begin{itemize}
\item $G^{DQN}_t$ ($G_t$ for the DQN): The $Q$-value $Q(S_{t+1}, a)$ used for the action selection
(red underline) and the $Q$-value $Q(S_{t+1}, a)$ used for the action evaluation (blue underline)
are determined by {\it the same neural network} with the weight vector $\theta_t$.
\\
\item $G^{doubleDQN}_t$ ($G_t$ for the double DQN): The $Q$-value used for action selection and
the $Q$-value used for action evaluation are determined by
{\it two different neural networks} with weight vectors $\theta_t$ and $\theta'_t$.
These networks are called {\it thr current} and {\it the target} network, respectively.
\end{itemize}
~\\
~\\
However, due to the slowly changing policy, estimates of the value of
the {\it current} and {\it target} neural networks are still too similar,
and this still causes a consistent overestimation, \cite{4}.

\subsection{Actor-critic architecture in the DDPG}

The DDPG was one of the first algorithms that tried to use the $Q$-learning technique
of DQN models for continuous action spaces. The DDPG stands for
deep deterministic policy gradient, \cite{7}.
In this case, we cannot use the maximization operator for the $Q$-values over all actions.
However, we can use a function approximator, a neural network representing the $Q$-values.
We presume that there exists a certain function $Q(s, a)$ that is differentiable
with respect to an action $a$. However, finding
\begin{equation*}
   \argmax_a(Q(S_t, a))
\end{equation*}
~\\
overall actions $a$ for the given state $S_t$ means that we must solve
the optimization problem at every time step. This is a very expensive task.
To overcome this obstacle, a group of researchers from DeepMind
used the actor-critic architecture, \cite{3}.
They used two neural networks: one, as in the DQN, was:
$Q$-network representing $Q$-values; the other one was
an actor function $\pi(s)$ supplying the action
that yields the maximum of the value function $Q(s, a)$. For the current
state $s = s_t$, we have
\begin{equation*}
     \pi(s_t) = a^* \text{, where } a^* = \argmax_a Q(s_t,a)
\end{equation*}
~\\
For any state $s$,
\begin{equation*}
     \max_a Q(s,a) \approx Q(s, \pi(s))
\end{equation*}

\subsection{A pair of independently trained critics in the TD3}
The actor-critic double DQN and DDPG suffer from overestimation.
In \cite[p.5]{4}, it was suggested that a failure can occur due to the interplay between
the actor and critic updates. Overestimation occurs when the policy is poor,
{\it ``and the policy will become poor if the value estimate itself is inaccurate''}.
In \cite{4}, the authors suggested using a pair of critics ($Q_{\theta_1}$, $Q_{\theta_2}$),
and taking the minimum value between them to limit overestimation.
It was originally supposed that there would also be $2$ actors ($\pi_1$, $\pi_2$) with cross updating:
$Q_{\theta_1}$ with $\pi_2$, $Q_{\theta_2}$ with $\pi_1$
(where $\pi_i$ is optimized with respect to $Q_{\theta_i}$).

\begin{equation*}
  \begin{split}
     & y_1 = r + \gamma Q_{\theta_2} (s', \pi_1(s'))  \\
     & y_2 = r + \gamma Q_{\theta_1} (s', \pi_1(s'))  \\
  \end{split}
\end{equation*}
~\\
According to \cite{4}, due to the required computational costs, a single actor can be used.
This restriction in the number of actors does not cause an additional bias.
The method with two critics outperforms many other algorithms including DDPG.

\section{\sc\bf Exploration as a major challenge of learning}
  \label{sec_explore}

\subsection{Why explore?}

In addition to overestimation, another problem is inherent in deep RL, and is no less difficult.
This is {\it exploration}. We cannot unconditionally believe in maximum values of a $Q$-table
or in an action function $\pi(s)$ supplying the best actions. Why not?
First, at the beginning stage of the training process, the corresponding neural network is still ``young and stupid'', and its maximum values or best actions are far from reality. Second, perhaps the maximum
values and the best actions are not those that will lead us to the optimal strategy after hard training.

\subsection{Exploration vs. exploitation}
Exploitation means, that the agent uses its accumulated knowledge to select the subsequent action.
In our case, this means that for a given state, the agent finds the following action that maximizes the $Q$-value. Exploration means that the subsequent action is selected randomly.
No rule determines which strategy, exploration or exploitation is better. The real goal is to find
a true balance between these two strategies. As we can see, the balance strategy changes
during the learning process.

\subsection{Exploration in the DQN and double DQN}

One way to ensure adequate exploration in the DQN and double DQN is to use the
annealing-greedy mechanism, \cite{12}.
For the first episodes, exploitation is selected with a small probability,
for example, $0.02$ (i.e., the action will be chosen very randomly), and the exploration
is selected with a probability of $0.98$. Starting at a certain number of episodes $M_\varepsilon$,
exploration is performed with a minimal probability $\varepsilon_m$.
For example, if $\varepsilon_m = 0.01$, exploitation is chosen with a probability of $0.99$.
The probability formula of exploration $\varepsilon$ can be realized as follows:

\begin{equation*}
   \varepsilon = \max(\frac{\varepsilon_m - 1}{M_\varepsilon}i + 1, \varepsilon_m),
\end{equation*}
~\\
where $i$ is the episode number. Let $M_\varepsilon = 100$, and $\varepsilon_m = 0.01$.
Then, the probability $\varepsilon$ of exploration is as follows:

\begin{table}[h]
  \centering
  \renewcommand{\arraystretch}{1.8}
  \begin{tabular} {|c|cccccccc|}
  \hline
    $i$ &   $0$ & $1$  & $2$ & $3$  & $\dots$ & $98$ & $99$ & $\ge 100$  \\
  \hline
    $\varepsilon$ &   $1$ & $0.9901$ & $0.9802$ & $0.9703$ & $\dots$  & $0.0298$ & $0.0199$ & $0.01$ \\
  \hline
\end{tabular}
  \vspace{2mm}
  \label{tab_explore}
\end{table}

\subsection{Exploration in the DDPG}

In RL models with continuous action spaces,
instead of the $\varepsilon$-greedy mechanism undirected exploration is applied.
This method is used in DDPG, proximal policy optimization (PPO) and other continuous control algorithms.
The authors of DDPG algorithm, \cite{3}, constructed an undirected exploration policy
$\pi’$ by adding noise sampled from a noise process $N$ to the actor policy $\pi(s)$:

\begin{equation*}
   \pi'(s_t) = \pi(s_t | \theta_t) + N,
\end{equation*}
~\\
where $N$ is the noise given by the Ornstein-Uhlenbeck, correlated noise process, \cite{8}.
In their TD3 paper, \cite{4}, the authors proposed using the classic Gaussian noise:
`` \dots {\it we use an off-policy exploration strategy, adding Gaussian noise
N(0; 0:1) to each action. Unlike the original implementation of DDPG,
we used uncorrelated noise for exploration as we found noise drawn
from the Ornstein-Uhlenbeck (Uhlenbeck \& Ornstein, 1930) process
offered no performance benefits.}''

The usual failure mode for DDPG is that the learned $Q$-function begins
to overestimate $Q$-values, and then the policy (actor function) leads to significant errors.

\subsection{Exploration in the TD3}

The name {\it TD3} stands for twin delayed deep deterministic algorithm. The TD3 algorithm retains the actor-critic architecture used in the DDPG, and adds $3$ new properties that greatly help it
to overcome overestimation:

\begin{itemize}
  \item The TD3 maintains {\it a pair of critics} $Q1$ and $Q2$ (hence the
  name “twin”) along with a single actor. At each time step, the TD3 uses the smaller
  of the two $Q$-values.
  ~\\
  \item The TD3 updates the policy (and target networks) less frequently than the
  $Q$-function updates (one policy update (actor) for every two $Q$-function (critic)
  updates).
  ~\\
  \item The TD3 adds {\it exploration noise} to the target action. {\it TD3} uses Gaussian noise,
  not Ornstein--Uhlenbeck noise as in the DDPG.
\end{itemize}

\section{\sc\bf PyBullet trained agents: Hopper, Walker2D and HalfCheetah}
  \label{sec_pybullet}

\begin{figure}[ht]
\centering
\includegraphics[scale=0.8]{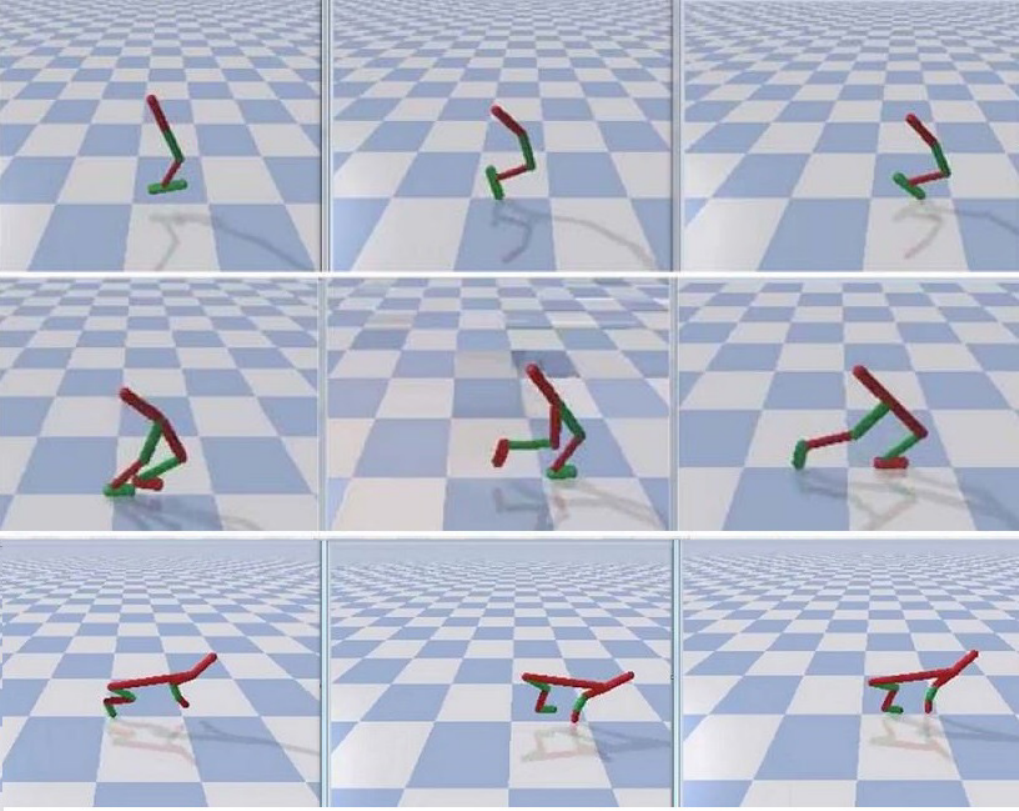}
\caption{\footnotesize Hopper, Walker2D and HalfCheetah trained agents}
\label{three_agents}
\end{figure}

PyBullet is a Python module for robotics and deep RL using PyBullet environments
are based on the Bullet Physics SDK, \cite{9, 15}. Let us look at
agents trained for HopperBulletEnv, Waker2DBulletEnv and HalfCheetahBulletEnv
which are typical PyBullet environments associated with articulated bodies, see Fig. \ref{three_agents}

~\\
~\\
~\\

\subsection{Exploration noise in trials with PyBullet Hopper}

The HopperBulletEnv environment is considered solved if the achieved score exceeds $2500$.
In TD3 experiments with the HopperBulletEnv environment, I got, among others,
the following training curves with $std = 0.1$ and $std = 0.3$:

\begin{figure}[ht]
\centering
\includegraphics[scale=0.8]{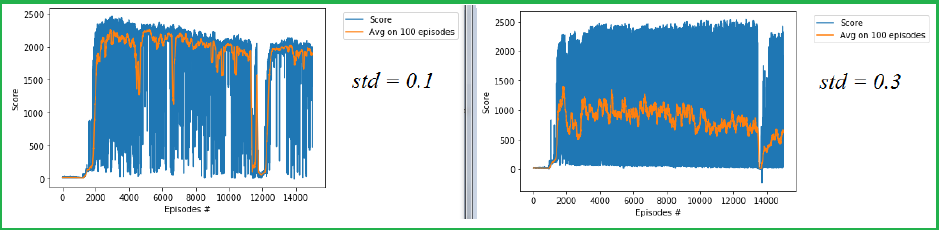}
\caption{\footnotesize Experiments with Hopper train agent}
\end{figure}
~\\
Here, $std$ is the standard deviation of exploration noise in the TD3.
In both trials, the threshold of $2500$ was not reached. However, I noticed the following features:

\begin{itemize}
  \item In the experiment with $std = 0.3$, there are a lot
  of values near $2500$ (but less than 2500)
  and at the same time, the average score decreased at all times. This is explained as follows:
  the number of small score values prevailed over the number of large score values,
  and the difference between these numbers increased.
  \\
  \item In the experiment with $std = 0.1$, the average score values reached
  large values but in general, the average scores decreased.
  The reason of this, as above, is that the number of
  small score values prevailed over the number of large score values.
  \\
  \item It seems that the prevalence of very small values was due to 
  too high noise standard deviations. Then, the decision was made 
  to reduce $std$ to $0.02$,
  this was sufficient for solving the environment.
\end{itemize}

\begin{figure}[h]
\centering
\includegraphics[scale=1.15]{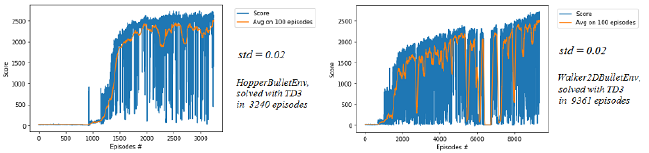}
\includegraphics[scale=0.95]{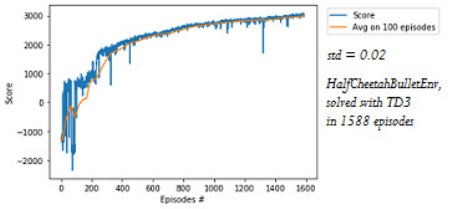}
\caption{\footnotesize Hopper,Walker2D and HalfCheetah are solved with TD3, see \cite{18}}
\end{figure}

\appendix

\section{\sc\bf Hill climbing algorithm with adaptive noise}

\subsection{Forerunner of tensors}

We illustrate the properties of the hill climbing algorithm, see \cite{6}, applied
to the {\it Cartpole} environment, \cite{11}.
Here, the network model is so simple that it does not use tensors
(no PyTorch and no Tensorflow). Only a simple matrix of size [4 x 2]
is used here, this is the forerunner of tensors.
The hill climbing algorithm seeks to maximize a target function $G_0$,
which in our particular case is the cumulative discounted reward:

\begin{equation*}
  G_0 = R_1 + \gamma R_2 + \gamma^2 R_3 + \dots = \sum_{k=0}^{\infty} {\gamma}^k R_{k+1}
\end{equation*}
~\\
where $\gamma$ is the {\it discount factor}, $0 < \gamma < 1$, and $R_k$ is the reward obtained
at the time step $k$ of the episode. The target function $G_0$ appears in Python as follows:
\begin{python}
   discounts = [gamma**i for i in range(len(rewards)+1)]
   ''' This is the 'cumulative discounted reward' or TD-target G_t '''
   Gt = sum([a*b for a,b in zip(discounts, rewards)])
\end{python}        

\subsection{Two Cartpole environmens}

What is $Cartpole$ ?
A pole is attached by a joint to a cart, which moves along a track.
The system is controlled by applying a force of +1 or -1 to the cart. The pendulum starts upright,
and the goal is to prevent it from falling over. A reward of +1 is provided for every timestep
that the pole remains upright. The episode ends when the pole $> 15$ degrees from a vertical orientation,
or the cart moves $> 2.4$ units from the center, \cite{12}.
The differences between Cartpole-v0 (resp. Cartpole-v1) are in two parameters:
the threshold = $195$ (resp. $475$) and the max number of episodes = $200$ (resp $500$).
Solving the Cartpole-v0 environment  (resp. Cartpole-v1) requires an average
total reward that exceeds the threshold for $100$ consecutive episodes.

For GitHub projects that solve both Cartpole-v0 and Cartpole-v1 environments  with
DQNs and double DQNs, see \cite{16,17}.

\subsection{Weight matrix in the hill climbing model}

Hill climbing is a simple gradient-free algorithm.
We try to climb to the top of a curve by only changing the arguments of the
target function $G_0$ using a certain {\it adaptive noise}.
The argument of $G_0$ is a weight matrix for determining the neural network
that underlies our model.

\subsection{Adaptive noise scale}

The adaptive noise scale for our model is realized as follows.
If the current value of the target function is better than the best value obtained
for the target function, we divide the noise scale by $2$, and the corresponding noise is added
to the weight matrix. If the current value of the target function is worse than
the best obtained value, we multiply the noise scale by $2$, and the corresponding noise is added
to the best obtained weight matrix value. In both cases, a noise scale
is added with some random factor different for each element of the matrix:

\begin{python}
  if Gt >= best_Gt: 
       ''' found better weights ==> decrease the noise: noise = noise/2'''
       best_w = policy.w
       noise_scale = max(1e-3, noise_scale/2)
       policy.w  += noise_scale * np.random.rand(*policy.w.shape)
  else:
      ''' no better weights, increase the noise: noise = noise/2'''     
      noise_scale = min(2, noise_scale * 2)
      policy.w  += best_w + noise_scale * np.random.rand(*policy.w.shape)      
\end{python}

The Cartpole-v0 environment is solved in $113$ episodes, and Cartpole-v1
is solved in $112$ episodes.  For more information on Cartpole-v0/Cartpole-v1 with
adaptive noise scaling, see the Jupyter notebooks \cite{16,17}.

\subsection{A more generic formula for the noise scale}

As seen above, the noise scale adaptively increases or decreases
depending on whether the target function is lower or higher than the best obtained value.
The noise scale in this algorithm is $2$.
In \cite{14}, the authors considered more generic formula:

\begin{equation*}
  \sigma_{k+1} = \left \{
    \begin{array}{ll}
       & \alpha \sigma_k \text{ if } d(\pi, \widetilde\pi) < \delta,  \\
       & \\
       & \frac{1}{\alpha} \sigma_k \text{ otherwise },
    \end{array}
    \right . \\
\end{equation*}
~\\
where $\alpha$ is a noise scale, $d$ is a certain distance measure between
the perturbed and nonperturbed policy, and $\delta$ is a threshold value.
In \cite[App. C]{14}, the authors considered the possible forms of the distance function
$d$ for the DQN, DDPG and TPRO algorithms.


\end{document}